\documentclass[preprint,12pt]{elsarticle}

\usepackage{amssymb}
\usepackage{bm}
\usepackage{xcolor}
\usepackage{algorithm}
\usepackage{algpseudocode}
\newcommand{\tab}{\hspace*{1.5em}}
\usepackage{pifont}
\usepackage{multirow}

\usepackage{amsmath}

\journal{Biomedical Signal Processing and Control}

\begin{document}

\begin{frontmatter}



\title{AdaptiveCDM: Source-Free Few-Shot Domain Adaptation for \underline{C}ell \underline{D}etection in \underline{M}icroscopic Images}



\author[label1]{Nimra Dilawar}
\author[label1]{Sara Nadeem}
\author[label2]{Javed Iqbal}
\author[label1]{Waqas Sultani}
\author[label1]{Mohsen Ali}


\affiliation[label1]{organization={Intelligent Machines Lab},
            addressline={Information Technology University }, 
            city={Lahore},
            country={Pakistan}}

\affiliation[label2]{organization={Computer Vision Researcher at Roll Camera Inc},
            city={Lahore},
            country={Pakistan}}


\begin{abstract}
Cross-domain cell detection for microscopic images suffers from performance degradation due to distribution shifts across imaging domains. 
Unsupervised Domain Adaptation (UDA) strategies, attempt to overcome domain sift without requiring annotated data from target. 
However, requirement of availability of annotated data from the source domain and large-size data from target domain are both challenging limitations for realistic scenarios. 
This is especially true in medical imaging, where privacy requirements might prevent access to annotated source data, and costly data acquisition restricts extensive sampling of the target domain.
To address these challenges, we propose AdaptiveCDM, a modular framework for \textit{Source-Free Few-Shot Domain Adaptive Object Detection} (SF-FSDAOD) setting, that adapts a pretrained source model using only few labeled target images without accessing source data.
AdaptiveCDM combines Resolution-Aware Augmentation (RAug) and Category-Aware Representation Learning (CARL). RAug alleviates the scarcity and class imbalance by augmenting instance balanced training examples, while preserving the scale fidelity and morphological properties of cellular structures. CARL enhances discriminative representation learning by encouraging class-consistent proposals, improving both localization and classification.
We also introduce two competitive baselines for proposed setting: Faster-FreeShot and MT-FreeShot. 
Our approach achieves 40.4/43.4 mAP@0.5 on M5 and 67.1/75.5 mAP@0.5 on Raabin-WBC under 2-/5-shot adaptation.
Despite using only a few labeled target images and no source data, AdaptiveCDM achieves competitive or superior performance compared with SOTA methods under their respective supervision settings.
Ablations and qualitative analyses further substantiate the contribution of each component and the effectiveness of AdaptiveCDM in low-data regimes. Code and models will be publicly available.
\end{abstract}

\begin{graphicalabstract}
\begin{figure*}[h]
    \centering
    \includegraphics[width=1.0\linewidth]{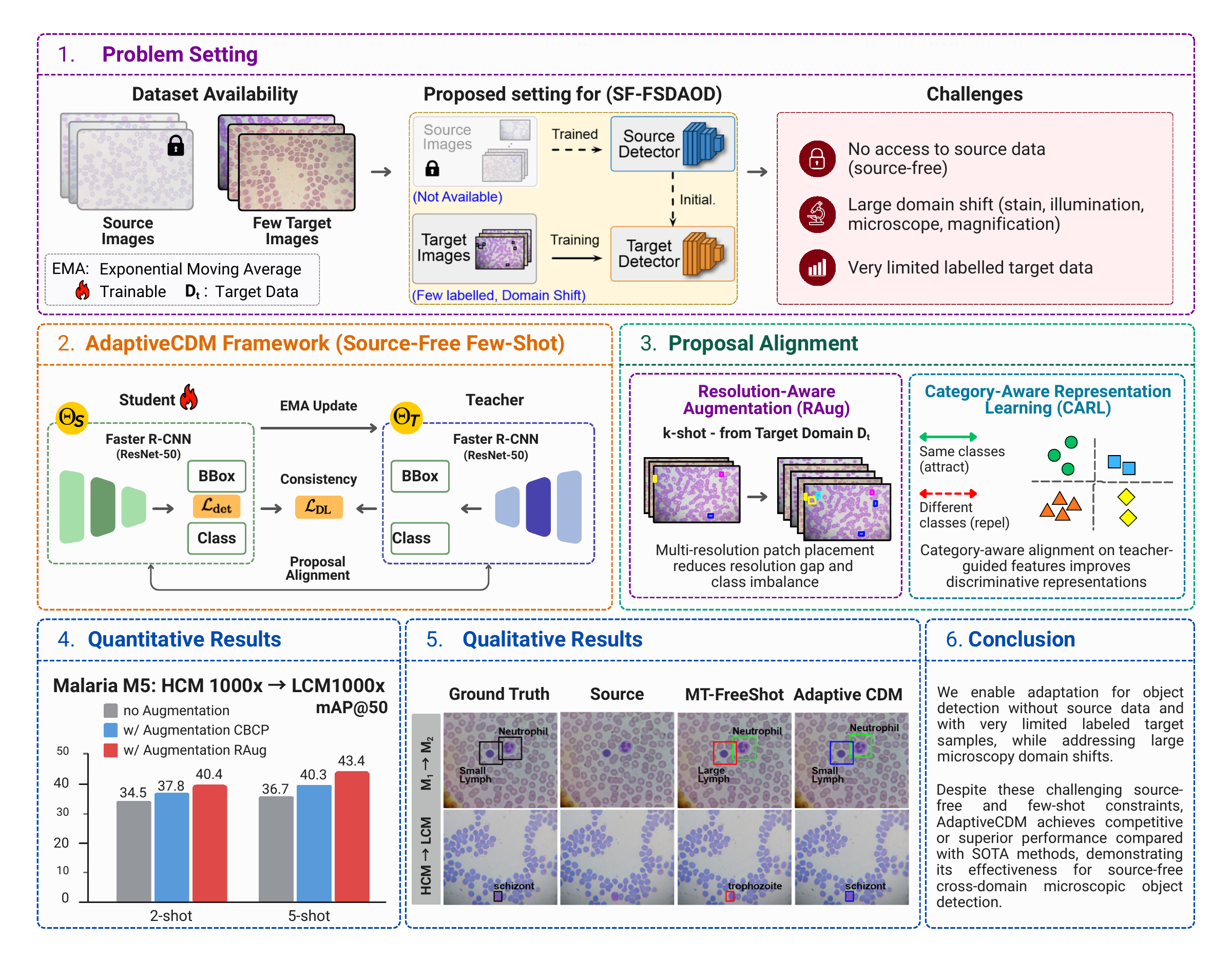}
    \label{fig:graphical_abstract}
\end{figure*}
\end{graphicalabstract}

\begin{highlights}

\item AdaptiveCDM enables source-free few-shot domain adaptation for microscopic object detection.

\item AdaptiveCDM adapts detectors without accessing source-domain images.

\item RAug mitigates resolution shift and class imbalance in few-shot microscopy.

\item CARL improves class discrimination using teacher-guided spatial cues.

\item AdaptiveCDM achieves strong results across two microscopy benchmarks.

\end{highlights}


\end{frontmatter}



\section{Introduction}
\label{sec1}
Blood examination is widely used for routine health assessment and the diagnosis of hematological and infectious diseases \cite{marin2024introduction,lynch1990peripheral}. 
Automated hematology analyzers have advanced, however microscopic examination of peripheral blood smear remain essential. Research shows that 42\% of cases still require this manual analysis \cite{asad2017utility}.  
Therefore, microscopic examination of peripheral blood smears remains an important component of hematological analysis. It enables the identification of different cell types and disease-associated cellular stages based on their visual characteristics and reduces analysis turnaround time \cite{katz2022remote}. The manual examination of blood-smear images is labor-intensive, time-consuming, and requires substantial expertise, especially when large numbers of samples must be examined or when abnormal cells are rare.

To reduce the workload of pathologists and enhance diagnostic efficiency, intelligent computer-aided diagnosis (CAD) methods have shown strong performance in detecting and localizing different targeted blood-cell in microscopic images \cite{abhishek2023automated,de2022new,lu2024blood,wu2026segmentation,chin2015intelligent,xue2019training,tellez2026deep}.
However, these approaches typically assume that the training/testing data are acquired under similar conditions and follow the similar distribution \cite{guan2021domain}. In practice this assumption is frequently violated due to variations in different microscopes like magnification, illumination and optical systems as well as differences in peripheral blood smear staining and image acquisition protocols \cite{wu2026segmentation}. Such discrepancies in source datasets (microscopic images from setup 1) cause performance degradation on target datasets (microscopic images from setup 2), commonly referred to as ``domain shift'' \cite{inayat2024few, sultani2022towards, rehman2024large}.



\begin{figure}[t]
    \centering
    \includegraphics[width=1.0\linewidth]{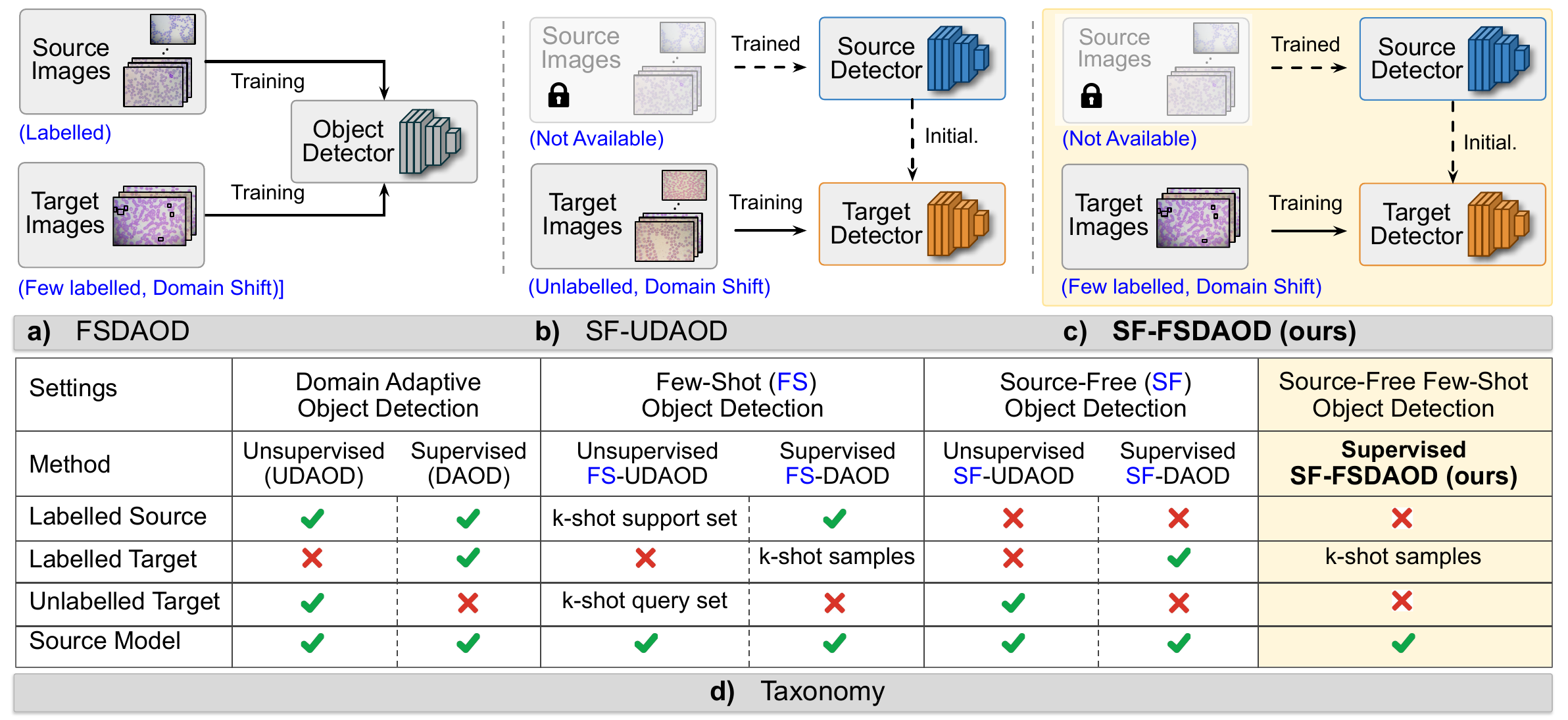}
    \caption{Comparison of Source-Free Few-Shot Domain Adaptive Object Detection (SF-FSDAOD) with other domain adaptation paradigms. \textbf{a)} FSDAOD (Few-Shot Domain Adaptive Object Detection): Access to source data and a few labeled target samples. \textbf{b)} SF-UDAOD (Source-Free Unsupervised Domain Adaptive Object Detection): The source data is inaccessible; only source model (trained on source data) and unlabeled target data are present. \textbf{c)} Proposed SF-FSDAOD (Source-Free Few-Shot Domain Adaptive Object Detection): The source data is inaccessible, and only \textit{k} target samples (labeled) are available alongside the source-trained model. \textbf{d)} Taxonomy illustrating the position of our proposed setting (SF-FSDAOD) for cross-domain object detection. }
    \label{fig:domain-adaptaion-setting}
\end{figure}

One of the common approaches to address domain shift in microscopic images is domain adaptation \cite{becker2014domain}. Supervised domain adaptation methods for object detection have shown strong performance in localizing and classifying cellular structures \cite{rehman2024large, sultani2022towards}. These methods rely on extensive annotated source and target domains data, making their deployment impractical in many clinical settings where data sharing is restricted and expert annotation is time-consuming and costly.
Another viable approach for addressing domain shift is unsupervised domain adaptation (UDA) \cite{iqbal2022fogadapt,xu2020exploring,munir2021ssal}. UDA methods leverage source data (labeled) together with target data (unlabeled).

Although UDA methods have demonstrated strong performance, their adoption for microscopic image analysis remains constrained by several practical limitations.
(1) Strict privacy and policy regulations often restrict access to sensitive source datasets \cite{zhou2025progressive}. (2) Collecting abundant target dataset is costly and knowledge-intensive \cite{inayat2024few, keaton2023celltranspose}. 
Moreover, since target images are unlabeled, rare classes may be underrepresented. This is a critical challenge for cell localization in microscopic imagery.

Source-domain patient data are often inaccessible due to privacy regulations and restrictions on data sharing
\cite{mitsuzumi2024understanding}, while collecting and annotating large target datasets requires substantial expert effort \cite{zhang2026reclaiming,zhang2026mind}. Therefore, the adaptation method should operate without having access to source data and minimize target data annotation requirements. 
It should leverage the knowledge encoded in a pretrained source model and adapt effectively using only a limited number of labeled target samples.
Practically a pretrained source model can be shared more readily, while a small number of target-domain microscopic images can be annotated by domain experts.
This motivates a source-free few-shot domain adaptation setting for object detection in which an existing detector is locally adapted to a new microscopic dataset using only few labeled target samples without requiring access to the original source images.
We systematically investigate this adaptation setting and propose \textbf{AdaptiveCDM} a source-free few-shot
domain Adaptation for Cell Detection in \textbf{M}icroscopic images) as illustrated in Figure \ref{fig:domain-adaptaion-setting}.

AdaptiveCDM addresses two major challenges: (1) severe class imbalance, where rare cell categories are substantially underrepresented in the few-shot target set, and (2) high visual similarity between foreground and background regions, as well as among different cellular categories and disease stages. Importantly, the proposed setting does not impose constraints on how the source model was originally trained, making the adaptation process flexible and broadly applicable.
The proposed setup closely reflects real-world medical imaging workflows, where institutions share pre-trained models rather than raw data due to privacy, and new data annotation is labor-intensive and requires expert domain knowledge. Despite its practical relevance, to the best of our knowledge, SF-FSDAOD remains largely unexplored and has not yet been systematically studied in the literature.



AdaptiveCDM integrates the source-free adaptation paradigm with few-shot object detection. To our knowledge, no established baselines exist for this combined setting. Consequently, we compare our approach against closely related state-of-the-art methods in (i) source-free domain adaptation and (ii) few-shot domain adaptive object detection. 
Most mainstream source-free object detection methods operate under the source-free unsupervised domain adaptation (SF-UDA) setting \cite{li2022source,vs2023instance,chu2023adversarial,varailhon2024source} and rely on abundant unlabeled target data.
Alternatively few-shot domain adaptive object detection (FSDAOD) methods \cite{gao2022acrofod,gao2023asyfod} assume complete access to source data with a few labeled target samples.

A limitation of these methods is their dependency on labeled source data, which makes them incompatible with the source-free setting.
Recent source-free few-shot domain adaptation (SF-FSDA) methods have shown progress in classification \cite{zhang2025multicentric,xu2024enhancing,yazdanpanah2022visual}
and segmentation \cite{zhou2025progressive, liu2025textual,vs2022target} tasks, yet remain largely unexplored for object detection. Sun et al.
\cite{sun2023sf} investigate SF-FSDA for object detection on two classes and is closely related to our work. However, it approaches the problem from a different perspective by relying on auxiliary textual descriptions for adaptation, introducing an additional modality and increasing computational overhead. In contrast, we address multi-class detection without using any metadata, making our approach more suitable for microscopic imagery.

\begin{figure*}[t]
    \centering
    \includegraphics[width=1.0\linewidth]{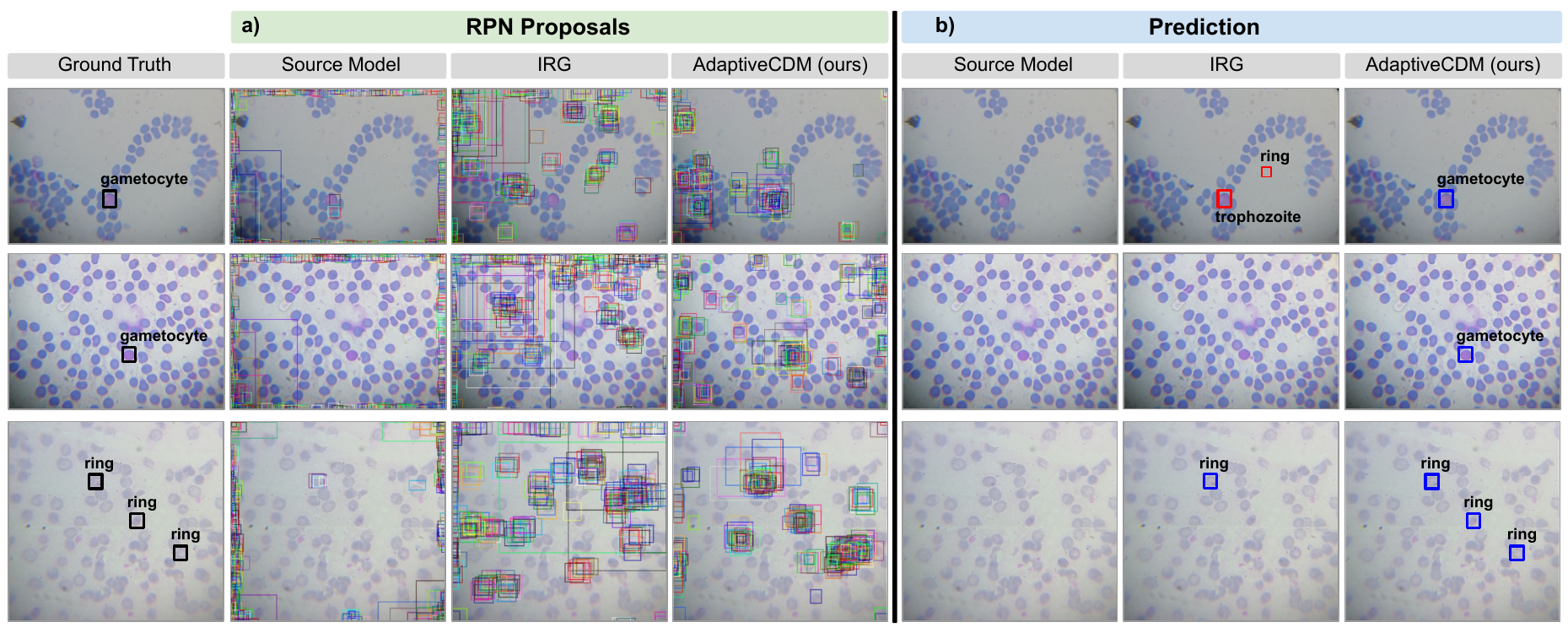}
    \caption{RPN proposal analysis on the M5 5-shot. (a) Region proposals generated by the source model, IRG, and AdaptiveCDM. (b) Corresponding final detections. AdaptiveCDM improves foreground–background discrimination and reduces false-positive detections.}
    \label{fig:rpn_proposals}  
\end{figure*}

\noindent
Our contributions are summarized as follows: 

\begin{itemize}
    \item We systematically investigate Source-Free Few-Shot Domain Adaptive Object Detection (SF-FSDAOD) for microscopic cell detection, a setting relevant to privacy-restricted healthcare applications, and propose AdaptiveCDM to adapt a pretrained source detector using only a few labeled target samples without accessing source data.
    \item AdaptiveCDM consists of two key components: 
     \begin{itemize} 
         \item Resolution-Aware augmentation strategy, `RAug', which alleviates the scarcity and class imbalance by augmenting instance balanced training examples, while preserving the scale fidelity and morphological properties of cellular structures.
         \item Category-Aware Representation Learning (CARL) to enhance discriminative representation learning by encouraging class-consistent proposals, improving both localization and classification. CARL aligns student features with teacher-guided class spatial cues using a joint similarity and dissimilarity objective to enforce intra-class compactness and inter-class separation (Figure \ref{fig:rpn_proposals})
     \end{itemize}
     \item In the absence of established baselines, we construct two competitive SF-FSDAOD baselines referred to as (i) Faster-FreeShot and (ii) MT-FreeShot.
     \item We compare AdaptiveCDM with related SOTA methods in (i) source-free domain adaptation and (ii) few-shot object detection. AdaptiveCDM consistently outperforms SF-UDA and FSDAOD baselines using only a few target samples without source data, with notable improvements on rare classes. 
\end{itemize}

\section{Related Work}
\label{sec:formatting}

\subsection{Unsupervised Domain Adaptation}
The pioneering work on unsupervised domain adaptation (UDA) object detection can be divided into four categories. \textbf{1)} The features-alignment methods include DAFaster \cite{chen2018domain}, SWDA \cite{saito2019strong}, HTCN \cite{chen2020harmonizing}, SSA-DA \cite{zhao2020bi}, ICRCCR \cite{xu2020exploring}, VDD \cite{wu2021vector}, SGA-S \cite{zhang2021self}, CST-DA \cite{zhao2020collaborative}, \cite{munir2021ssal}, and DBGL \cite{chen2021dual} align feature embeddings at image-level or instance-level using both labeled source and unlabeled target domain data. \textbf{2)} Self-training based methods, NL \cite{khodabandeh2019robust}, CDG \cite{li2021category}, SSAL \cite{munir2021ssal}, STMDA-Retinanet \cite{pasqualino2022multi} acquire pseudo instance-level annotations to incorporate domain invariant information. \textbf{3)} Based on image-translation strategy \cite{zhu2017unpaired}, other methods DM \cite{kim2019diversify}, AFAN \cite{wang2021afan}, UMT \cite{deng2021unbiased}, perform adaptation using both original target domain images and source-styled target images with the source domain images. \textbf{4)} Mean-Teacher framework-based methods include NL \cite{khodabandeh2019robust}, ICR-CCR \cite{xu2020exploring}, MTOR \cite{cai2019exploring}, UMT \cite{deng2021unbiased}, which utilizes an auxiliary multi-class classifier, an identical network to a task-specific detector.
UDA methods achieve promising results, but there is a need to access labeled source and unlabeled target domain data during adaptation.

\subsection{Source-free Domain Adaptation (SFDA)}
Source-Free Domain Adaptation (SFDA) has gained significant attention in recent years, particularly for addressing classification problems \cite{li2020model, liang2020we, cai2025adaptive}. Extensive research has been conducted on SFDA for image classification, employing various strategies such as source-style image generation SFIT \cite{hou2021visualizing}, source-style feature generation VDM-DA \cite{tian2021vdm}, and target-style image generation SDDA \cite{kurmi2021domain}. Similarly, other methods that use self-training-based approaches, such as SHOT \cite{liang2020we} and SHOT++ \cite{liang2021source}, generate pseudo-labels by leveraging class centroids. SFDA has also been explored for more complex tasks such as object detection and semantic segmentation \cite{yu2023source, lo2023spatio}.
Most SFDA based object detection methods follow an unsupervised domain adaptation (UDA) self-training setting in the absence of source domain data. LODS \cite{li2022source} enhances the style of target domain images using a style-transfer module and performs graph-based alignment at the image and instance level features. IRG \cite{vs2023instance} learns contrast between target objects through a graph convolutional network. SFOD \cite{li2021free} exploits the entropy measure to determine an optimal confidence threshold for pseudo-labels generation. The mean-teacher-based framework $\mathrm{A^2SFOD}$ \cite{chu2023adversarial} obtains refined pseudo-labels by aligning the source-similar and source-dissimilar target features space using adversarial learning.
\subsection{Few-Shot Domain Adaptation (FSDA)}
Several few-shot domain adaptation (FSDA) methods, such as VFA \cite{han2023few}, FDP \cite{han2023few}, address domain shift challenges using labeled source data and a few labeled target samples by leveraging meta-learning-based approaches. These methods adapt the few-shot object detection (FS-OD) setting to FSDA by treating base class samples as source data and novel samples as target data. Subsequent approaches extended this idea to the unsupervised domain adaptation (UDA) setting. For instance, AcroFOD \cite{gao2022acrofod} applies weak and strong domain-aware augmentations at both the image and instance levels to enhance the diversity of target domain samples. Building on this, AsyFOD \cite{gao2023asyfod} addresses the over-adaptation issue caused by imbalanced data distributions, generalizing the work of \cite{gao2022acrofod}.
In the medical domain, \cite{su2021few} proposed a few-shot domain adaptation (FSDA) method for anomaly detection in fMRI brain images, leveraging only a limited number of labeled samples. Similarly, \cite{keaton2023celltranspose} proposed a cellular instance segmentation method that employs contrastive losses to address covariate shifts while minimizing retraining time.

\subsection{Source-Free Few-Shot Domain Adaptation (SF-FSDA)}
Growing concerns around data privacy and limited annotation budgets have brought attention to the SF-FSDA setting in image classification.
\cite{xu2024enhancing} proposed an information maximization (IM-DCL) approach for implicit alignment, which learns the query set through a transduction mechanism. VDB \cite{yazdanpanah2022visual} is a test-time batch normalization method that aligns the source and target distributions without updating the model weights. However, object detection under SF-FSDA remains insufficiently studied. Han et al. \cite{sun2023sf} proposes a method in the SF-FSDA setting for object detection, which synthesizes target-like images using a few target samples and depends on additional supervision through textual descriptions. The proposed method is limited to only two object classes, and the code for reproducibility is not publicly available. 
Our work fills this gap by investigating multiclass object detection under SF-FSDA for microscopic imagery, with an emphasis on practical, privacy-preserving adaptation.


\begin{figure*}[t]
  \begin{minipage}[t]{\linewidth}
    \centering
    \includegraphics[width=1.0\linewidth]{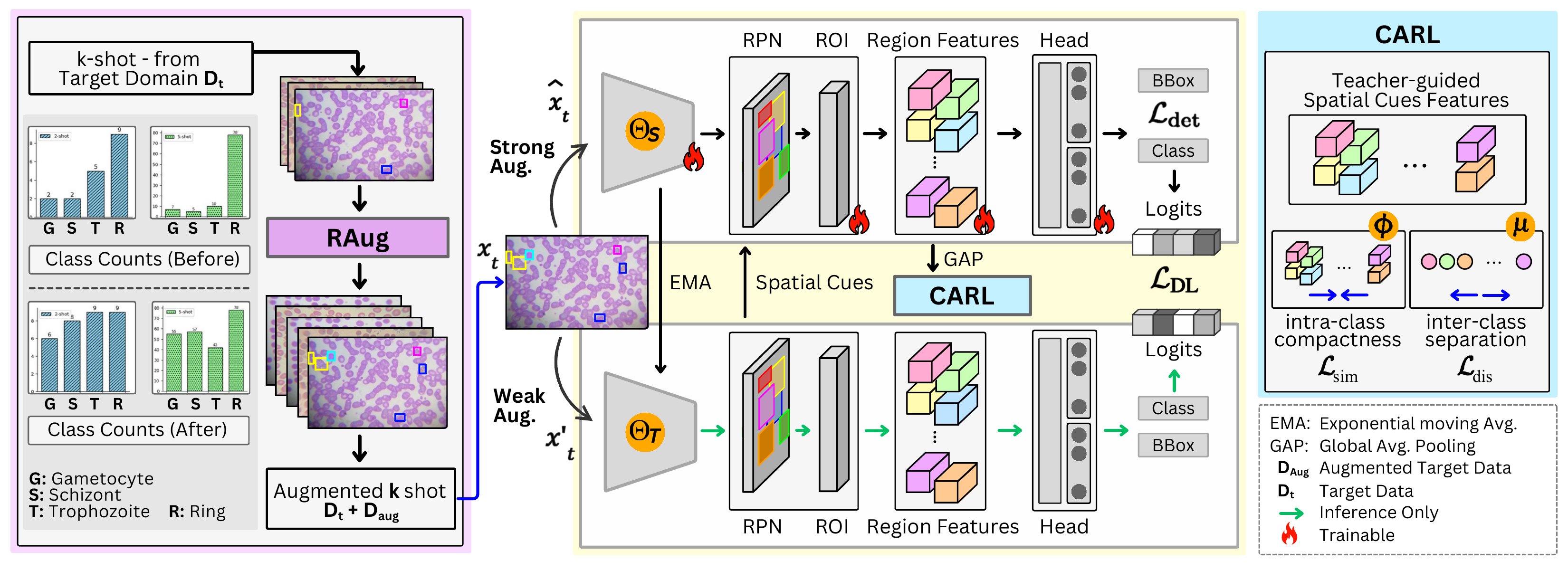}
    \caption{\textbf{Proposed Method}: AdaptiveCDM adopts a student–teacher framework based on a two-stage detector \cite{ren2016faster} with a backbone, RPN, and ROI Align. Both models are initialized from the source-trained network. Adaptation to the target domain uses only k-shot samples from {$D_t$}, with augmented samples precomputed for training. The teacher is updated via an Exponential Moving Average (EMA) of the student weights. }
    \label{fig:main_flow}
  \end{minipage}
\end{figure*}

\section{Method}
\label{subsec:method}

This section outlines the overall pipeline of our proposed framework, as shown in Figure \ref{fig:main_flow}. We first describe the preliminary setup in Section \ref{subsec:preliminaries}, followed by the competitive baselines in Section \ref{subsec:Baselines}. Finally, Section \ref{subsec:AdaptiveMCD} presents our framework AdaptiveCD, which integrates Resolution-Aware Augmentation (RAug) and Category-Aware Representation Learning (CARL) to address source-free and few-shot domain adaptation challenges.

\subsection{Preliminaries}  
\label{subsec:preliminaries}
Let $\bm{D_s}=\{(x_i^s,y_i^s)\}_{i=1}^{N_s}$ be a labeled source dataset of size $N_s$, 
where $x_i^s \in \mathbb{R}^{H_i \times W_i \times 3}$ is an RGB image and 
$y_i^s=\{c_{i,j}^s, b_{i,j}^s\}_{j=1}^{O_i}$ denote the class and bounding-box annotations for $O_i$ objects. 
Let $\bm{D_t}=\{(x_i^t,y_i^t)\}_{i=1}^{N_t}$ be a $\textbf{k}$-shot target dataset with $\textbf{k}$ labeled images per class. An individual image may contain instances from multiple classes, so the total number of unique target images is not necessarily equal to $kC$.
The target dataset is constructed using k labeled samples following \cite{inayat2024few} with its overall size being substantially smaller than that of the source dataset, i.e. $N_t << N_s$.
Both domains share the same label space $c \in \{1,\dots,C\}$. 
Unlike conventional FSDAOD and UDA settings, our method assumes access only to a source model and few labeled target samples $\bm{D_t}$.

\subsection{Proposed Baselines for SF-FSDAOD}
\label{subsec:Baselines}
In microscopic images, precise localization and accurate classification are more important than processing speed. 
Hence, we adapt Faster R-CNN (FRCNN) at the proposal level to handle domain shift in region localization.
We introduce two competitive baselines by fine-tuning a source-trained model, using only a few labeled target images, \bm{$D_t$}. Specifically, we introduce \textbf{Faster-FreeShot} and \textbf{MT-FreeShot} for microscopic cell detection.



\subsubsection{Faster-FreeShot.}
Let $\mathcal{F}_{\text{det}}$ denote the FRCNN-based detector, initially trained on source images $\bm{D_s}$, called source model. In the adaptation phase, $\mathcal{F}_{det}$ is fine-tuned using few labeled samples $\bm{D_t}$ in supervised setting. The fine-tuning process is guided by a task-specific detection loss including classification and localization 

\begin{equation}
\mathcal{L}_{\text{det}} =
\mathcal{L}^{rpn}_{cls} + \mathcal{L}^{rpn}_{loc} + \mathcal{L}^{roi}_{cls} + \mathcal{L}^{roi}_{loc},
\label{eqn:L_det}
\end{equation}


where $\mathcal{L}^{rpn}_{cls}$ and $\mathcal{L}^{rpn}_{loc}$ are from the region proposal network (RPN) which gives region proposals, and $\mathcal{L}^{roi}_{cls}$ and $\mathcal{L}^{roi}_{loc}$ are the corresponding losses from the region of interest (ROI) head. Given image $x$ ROI head produces Z detections $\{P(\hat{c_n}\mid x), {\hat{b_n}}\}_{n=1}^{Z}$ , where $\hat{c_n}$ denotes the predicted class predictions and $\hat{b_n}$ denotes the predicted bounding box for the $n^{th}$ proposal, respectively. 
Given the shared label space between the source and target domains, and the availability of only a few labeled target samples, this fine-tuning strategy enables quick adaptation of the source model while retaining knowledge acquired during the source domain training.

\subsubsection{MT-FreeShot.} 
\label{sec:mt-freeshot}
Inspired by the generalization ability of the Mean Teacher (MT) framework~\cite{tarvainen2017mean,vs2023instance}, we introduce a source-free baseline, MT-FreeShot, tailored for SF-FSDAOD. 
Unlike standard fine-tuning in Faster-FreeShot, which is prone to overfitting due to limited labeled target data, MT-FreeShot employs consistency regularization between a student network $\Theta_{S}$ and a temporally averaged teacher $\Theta_{T}$, both initialized from the source model $\mathcal{F}_{det}$. 
By aligning their predictions, it stabilizes training, reduces noisy student updates, and improves generalization.

Given a target image \bm{$x_t$}, we obtain two augmented views \bm{${x'}_t$} and \bm{$\hat{x}_t$} by applying weak and strong augmentations \cite{vs2023instance}, which are then fed to $\Theta_{T}$ and $\Theta_{S}$, respectively. 
The consistency between $\Theta_{T}$ and  $\Theta_{S}$ is enforced using the Kullback–Leibler divergence loss $\mathcal{L}_{\text{DL}}$, applied to their respective output logits $Z_{T}$ and $Z_{S}$.

\begin{equation}
\mathcal{L}_{\text{DL}} =
\mathcal{KL}(\sigma(Z_{S}), \sigma(Z_{T})),
\label{eqn:L_DL}
\end{equation}where $\sigma$ denotes the softmax operation. The student network is trained by minimizing $\mathcal{L}_{\text{det}}$ 
while the teacher network weights are updated using an Exponential Moving Average (EMA) strategy as 

\begin{equation}
\Theta_{T} = \eta \cdot \Theta_{T} + (1-\eta)\cdot\Theta_{S},
\label{eqn:ema}
\end{equation} with $\eta=0.9$. Since the source images are not accessible during the adaptation, the teacher model remains more stable and helps retain source knowledge, preventing it from being forgotten.
The overall objective function for MT-FreeShot is 

\begin{equation}\mathcal{L}_{MT} = \mathcal{L}_{\text{det}} + \mathcal{L}_{\text{DL}}.
\end{equation}

\subsection{AdaptiveCDM: SF-FSDAOD for
Microscopic Cell Detection} 
\label{subsec:AdaptiveMCD}

AdaptiveCDM extends MT-FreeShot with two components: (i) Resolution-Aware Augmentation (RAug) and (ii) Category-Aware Representation Learning (CARL).

\subsubsection{Resolution-Aware Augmentation (RAug): }
\label{subsec:RAug}

Let $I_1 \in \mathbb{R}^{h_1\times w_1\times 3}$ be an image containing instances of class $p$, and $I_2 \in \mathbb{R}^{h_2\times w_2\times 3}$ denote a target image containing an underrepresented instance of the same class. We copy annotated cell patches $p_k$ from $I_1$ to empty regions in $I_2$. To preserve size consistency across images, we leverage the relation between the spatial resolution of $I_1$ and $I_2$. We select $k^{th}$ cell patch $p_k$ having resolution $(w_k^p, h_k^p)$ and place it in $I_2$, keeping the resolution intact if both images have the same spatial dimensions. If this condition is not satisfied, $p_k$ is resized to align with $I_2$ resolution while keeping its aspect ratio ${P}_{ratio} = h_k^p / w_k^p$ intact. $(\hat{w}_k^p, \hat{h}_k^p)$ in the adjusted resolution. The resized patches are pasted into $I_2$ after applying random color intensity variation and Gaussian blur.
\begin{algorithm}[H]
\caption{AdaptiveCDM training algorithm}
\label{algo:AdaptiveMCD}
\begin{algorithmic}[1]
\State \textbf{Inputs:} 
    \newline\tab Source model: {$\mathcal{F}_{det}$}, 
    \newline\tab $k$-shot labeled target: $\mathcal{D}_t$

\State \textbf{Define:}
\newline\tab Teacher network $\Theta_{te}$
\newline\tab Student network $\Theta_{st}$
\newline\tab Weak augmentation $\text{w}_{aug}(\cdot)$
\newline\tab Strong augmentation $\text{s}_{aug}(\cdot)$
\newline\tab Resolution-Aware Augmentation $\text{RAug}(\cdot)$
\State \textbf{Output:} Adapted $\Theta_{te}$

\State \textbf{Augmentation:} 
\newline\tab$\mathcal{D}_{aug}$ =  $\text{RAug}(\mathcal{D}_t)$
\newline\tab$\mathcal{D}_{t}$ =  $\mathcal{D}_{t}$ + $\mathcal{D}_{aug}$ 

\State \textbf{Adaptation pipeline:} 
\State \tab\textbf{Initialize:} $\Theta_{te} \leftarrow \mathcal{F}_{det}$, $\Theta_{st} \leftarrow \mathcal{F}_{det}$ 

\State \tab \textbf{for} each ${x}^t$ in ${\mathcal{D}}_t$ \textbf{do}
        \State \tab\tab ${\hat{x}}= \text{s}_{aug}({x}^t)$ , ${{x^\prime}}= \text{w}_{aug}({x}^t)$
        \State \tab\tab $\hat{\textbf{y}}^t$ , $Z_{te}$ = $\Theta_{te} (x^\prime )$ \Comment $\hat{\textbf{y}}^t=\{({b^\phi_r}~,~{c^\phi_r})\}_{r=1}^{N_r}$
        \State \tab\tab  $Z_{st}$ = $\Theta_{st} (\hat{x}, \hat{\textbf{y}}^t)$
        \State \tab\tab \textbf{Compute the loss: }
        \State \tab\tab\tab  $\mathcal{L}_{\text{det}}$ for $\Theta_{st}$ using Eq. \ref{eqn:L_det}    \Comment{Supervised Loss} 
        \State \tab\tab\tab  $\mathcal{L}_{\text{DL}}$ on $Z_{st}$ and  $Z_{te}$ using Eq. \ref{eqn:L_DL}  \Comment{Consistency Loss}
        \State \tab\tab\tab  $\mathcal{L}_{\text{sim}}$ and $\mathcal{L}_{\text{dis}}$ using Eq, \ref{eqn:l_contrast_sim} and Eq. \ref{eqn:l_contrast_dis}         \Comment{CL Loss}
        \State \tab\tab\tab Update $\Theta_{st}$ using $\mathcal{L}_{\text{CARL}}$ = 
        $\mathcal{L}_{\text{MT}}$ + 
        $\beta \mathcal{L}_{\text{dis}}$ + 
        $\alpha \mathcal{L}_{\text{sim}}$
        \State\tab\tab Update $\Theta_{te} \leftarrow \text{EMA}(\Theta_{st})$ using Eq. \ref{eqn:ema}
\State \hspace*{1.5em} \textbf{end for}

\end{algorithmic}
\end{algorithm}
\begin{equation}
(\hat{w}_k^p, \hat{h}_k^p) =
\begin{cases}
\label{eqn:raug}
    \left(\mathcal{P}_{\text{ratio}} \times \frac{h_2}{h_1} \times h_k^p, \quad \frac{h_2}{h_1} \times h_k^p \right) 
    & \text{if}~~~~ h_2 > w_2\\
    \left( \frac{w_2}{w_1} \times w_k^p, \frac{w_2}{w_1} \times w_k^p \times \mathcal{\text{P}}_{\text{ratio}} \right), 
    & \text{otherwise}\\
\end{cases}
\end{equation}

Microscopic imagery is inherently imbalanced \cite{rehman2024large}, where rare classes co-occur alongside common ones, which further aggravates imbalance in few-shot settings.
Although CutMix-based methods \cite{gao2022acrofod,gao2023asyfod,inayat2024few} attempt to address this, they overlook spatial resolution variations.
As observed in the microscopic benchmark \cite{sultani2022towards}, substantial resolution variability arises due to heterogeneous imaging devices and acquisition settings, making such augmentations suboptimal. To address this, RAug mitigates data scarcity and class imbalance by generating instance-balanced samples while preserving scale and morphology.




\subsubsection{Category-Aware Representation Learning (CARL). }
\label{subsec:CARL}
The core idea of CARL is to enhance foreground–background discrimination using augmented variants of foreground classes, thereby improving RPN localization and reducing false positives. 
While prior methods~\cite{cao2023contrastive,ma2025source} align teacher–student features or output probabilities, our proposed loss aligns augmented and non-augmented student features from different spatial locations with teacher-guided class-level spatial cues (region proposals).
We select $R=300$ teacher-guided spatial cues $\hat{y}$ with top objectness scores from the teacher’s RPN and construct class-wise cues using their labels and boxes. 
Although reliable, these cues may contain minor inaccuracies or miss objects; therefore, we merge them with ground-truth boxes $y$ to obtain $\tilde{y}=\{\hat{y}\cup y\}$. 
For each cue $(b,c)\in\tilde{y}$, we extract student backbone features $\boldsymbol{\phi}$ and apply global average pooling (GAP). 
To enforce intra-class compactness and inter-class separation, we define
\begin{equation}
\mathcal{L}_{\text{sim}}=\frac{1}{C}\sum_{c=1}^{C}\mathbb{E}_{j\neq k}\!\left[D(\phi_{c,j},\phi_{c,k})^2\right],
\label{eqn:l_contrast_sim}
\end{equation}

\begin{equation}
\mathcal{L}_{\text{dis}}=\frac{1}{\binom{C}{2}}\mathbb{E}_{c\neq c'}\!\left[\max\!\left(D(\mu_c,\mu_{c'})^2-m,0\right)\right].
\label{eqn:l_contrast_dis}
\end{equation}

\normalsize

$D( \cdot , \cdot )$ is the cosine distance, $\boldsymbol{\phi_{c,j}}$ and $\boldsymbol{\phi_{c,k}}$ denotes the feature embeddings of the $j^{th}$ and $k^{th}$ proposal of class $c$ respectively and $m = 1$ is the margin. 
$\boldsymbol{\mu_c}$ denotes the mean embedding of class $c$.
The overall loss is defined as 

\begin{equation}
\mathcal{L}_{CARL} = \mathcal{L}_{MT}+ \alpha \, \mathcal{L}_{\text{sim}} + \beta \, \mathcal{L}_{\text{dis}} , 
\label{eqn:CARL}
\end{equation}

where $\alpha=1$, $\beta=1$ are weighting coefficients, while $\mathcal{L}_{\text{sim}}$ and $\mathcal{L}_{\text{dis}} $ serve as regularization terms.

CARL improves feature separability and RPN foreground–background discrimination without source data. In contrast, most traditional contrastive methods rely on source-domain information leveraging source memory bank~\cite{lang2023class} or auxiliary predictor alignment~\cite{zhao2022task}, which are inapplicable in the source-free setting. The execution flow of the AdaptiveCDM framework is summarized in the pseudo-algorithm presented in Algorithm \ref{algo:AdaptiveMCD}.

\begin{table*}[t]
\centering
\scriptsize
\setlength{\tabcolsep}{5pt}     
\renewcommand{\arraystretch}{0.9} 
\fontsize{8.0pt}{12.0pt}\selectfont
\caption{Detection results of AdaptiveCDM on M5 and Raabin-WBC. Class-wise AP@0.5(\%) and overall mAP@0.5(\%) are reported. Mean and standard deviation over three runs are shown, with best results in \textbf{bold}.}
\label{tab:main_results}

\begin{tabular}{l | cc | cc | cc | cc | cc}
\noalign{\hrule height 1pt}

\multicolumn{11}{c}{M5 (HCM1000x $\rightarrow$ LCM1000x)}\\
\hline
\textbf{Method} & \multicolumn{2}{c}{\textbf{gamet.}} & \multicolumn{2}{c}{\textbf{schi.}} & \multicolumn{2}{c}{\textbf{troph.}} & \multicolumn{2}{c|}{\textbf{ring}} & \multicolumn{2}{c}{\textbf{mAP$_{50}$}} \\
\hline
Source~\cite{ren2016faster} & \multicolumn{2}{c}{0.0} & \multicolumn{2}{c}{0.0} & \multicolumn{2}{c}{43.3} & \multicolumn{2}{c|}{16.8} & \multicolumn{2}{c}{15.0} \\
\hline
Shots & 2 & 5 & 2 & 5 & 2 & 5 & 2 & 5 & 2 & 5 \\
\hline
Faster-FreeShot & 37.0 & 30.0 & 11.5 & 12.6 & 59.7 & 52.0 & 35.5 & 32.3 & \underline{35.9}$_{\pm2.8}$ & 31.7$_{\pm5.0}$ \\
MT-FreeShot & 41.0 & 43.1 & 0.0 & 2.6 & 63.7 & 64.0 & 33.3 & 37.3 & 34.5$_{\pm0.3}$ & \underline{36.7}$_{\pm1.5}$ \\
AdaptiveCDM & 52.9 & 56.9 & 7.4 & 15.2 & 64.7 & 64.0 & 36.5 & 37.4 & \textbf{40.4}$_{\pm0.9}$ & \textbf{43.4}$_{\pm0.4}$ \\
\hline
Oracle~\cite{ren2016faster} & 74.4 & & 0.0 & & 80.5 & & 64.2 & & 54.8 & \\
\noalign{\hrule height 1pt}
\multicolumn{11}{c}{Raabin-WBC (M1 $\rightarrow$ M2)}\\
\hline
\textbf{Method} & \multicolumn{2}{c}{\textbf{l-lymph}} & \multicolumn{2}{c}{\textbf{neutro.}} & \multicolumn{2}{c}{\textbf{s-lymph}} & \multicolumn{2}{c|}{\textbf{monoc.}} & \multicolumn{2}{c}{\textbf{mAP$_{50}$}} \\
\hline
Source~\cite{ren2016faster} & \multicolumn{2}{c}{30.4} & \multicolumn{2}{c}{40.5} & \multicolumn{2}{c}{71.9} & \multicolumn{2}{c}{56.8} & \multicolumn{2}{c}{49.9} \\
\hline
Shots & 2 & 5 & 2 & 5 & 2 & 5 & 2 & 5 & 2 & 5 \\
\hline
Faster-FreeShot & 22.1 & 37.6 & 80.2 & 85.6 & 93.9 & 93.9 & 75.9 & 74.2 & \textbf{68.0}$_{\pm0.5}$ & \underline{72.8}$_{\pm1.4}$ \\
MT-FreeShot & 16.9 & 30.7 & 61.8 & 70.7 & 90.7 & 91.7 & 73.2 & 72.7 & 60.6$_{\pm1.3}$ & 66.5$_{\pm1.1}$ \\
AdaptiveCDM & 24.3 & 51.2 & 77.8 & 81.6 & 93.3 & 93.6 & 73.0 & 75.7 & \underline{67.1}$_{\pm0.1}$ & \textbf{75.5}$_{\pm0.8}$ \\
\hline
Oracle~\cite{ren2016faster} & \multicolumn{2}{c}{93.0} & \multicolumn{2}{c}{92.9} & \multicolumn{2}{c}{96.4} & \multicolumn{2}{c}{81.4} & \multicolumn{2}{c}{90.9} \\
\hline
\end{tabular}

\end{table*}

\renewcommand{\arraystretch}{1}
\setlength{\tabcolsep}{2.1pt}         

\begin{table*}[t]
\centering

\setlength{\tabcolsep}{4pt}     
\renewcommand{\arraystretch}{1} 

\fontsize{8.0pt}{12.0pt}\selectfont
\caption{Detection comparison of AdaptiveCDM on M5 and Raabin-WBC. Class-wise and overall mAP@0.5(\%) are reported. SF-UDAOD utilizes the entire unlabeled target, while AdaptiveCDM uses only 5-shot labeled samples.}
\label{tab:M5_results_few_shot_comparison_with_SFUDA}
\begin{tabular}{ @{\hspace{1pt}} l 
| @{\hspace{1pt}} l 
| @{\hspace{1pt}} l | l c c c c | c }


\noalign{\hrule height 1pt}
\multicolumn{9}{c}{Raabin-WBC (M1 $\rightarrow$ M2)} \\
\hline

\multicolumn{1}{c}{} & \textbf{Method} & \textbf{arch.} & \multicolumn{1}{c|}{\textbf{t-imgs}} &
\multicolumn{1}{c}{\textbf{l-lymph}} &
\multicolumn{1}{c}{\textbf{neutr.}} &
\multicolumn{1}{c}{\textbf{s-lymph.}} &
\multicolumn{1}{c|}{\textbf{mono.}} &
\multicolumn{1}{c}{\textbf{mAP$_{50}$}} \\
\hline
\multirow{3}{*}{\textcolor{blue}{SF}\text{-UDA}}

& IRG \cite{vs2023instance} & FR50 & \multicolumn{1}{c|}{2033} &
\multicolumn{1}{c}{11.3} &
\multicolumn{1}{c}{53.4} &
\multicolumn{1}{c}{\underline{92.3}} &
\multicolumn{1}{c|}{\underline{60.0}} &
\multicolumn{1}{c}{\underline{54.2}} \\
& LODS \cite{li2022source} & FR101 & \multicolumn{1}{c|}{2033} &
\multicolumn{1}{c}{17.3} & 
\multicolumn{1}{c}{55.8} &
\multicolumn{1}{c}{73.2} &
\multicolumn{1}{c|}{0.0} &
\multicolumn{1}{c}{36.6} \\
& AASFOD \cite{chu2023adversarial} & FR50 & \multicolumn{1}{c|}{2033} &
\multicolumn{1}{c}{47.1} &
\multicolumn{1}{c}{38.4} &
\multicolumn{1}{c}{67.2} & 
\multicolumn{1}{c|}{7.2}   &
\multicolumn{1}{c}{40.0} \\
& SF-YOLO \cite{varailhon2024source} & yolov5 & \multicolumn{1}{c|}{2033} &

\multicolumn{1}{c}{37.1} & 
\multicolumn{1}{c}{\textbf{82.8}} &
\multicolumn{1}{c}{21.0} &
\multicolumn{1}{c|}{9.7} &
\multicolumn{1}{c}{37.7} \\
\hline
\multirow{1}{*}{\textcolor{blue}{SF}\text{-FSDA}}
&\textbf{AdaptiveCDM}& FR50 & \multicolumn{1}{c|}{\textbf{20}} &
\multicolumn{1}{c}{\textbf{51.2}} &
\multicolumn{1}{c}{\underline{81.6}} &
\multicolumn{1}{c}{\textbf{93.6}} & 
\multicolumn{1}{c|}{\textbf{75.7}} &
\multicolumn{1}{c}{\textbf{75.5}} \\

\noalign{\hrule height 1pt}
\multicolumn{9}{c}{M5 (HCM1000x $\rightarrow$ LCM1000x)}\\
\hline

\multicolumn{1}{c}{} & \textbf{Method} & \textbf{arch.} & \multicolumn{1}{c|}{\textbf{t-imgs}} &
\multicolumn{1}{c}{\textbf{gamet.}} & 
\multicolumn{1}{c}{\textbf{schi.}} & 
\multicolumn{1}{c}{\textbf{troph.}} & 
\multicolumn{1}{c|}{\textbf{ring}} &
\multicolumn{1}{c}{\textbf{mAP$_{50}$}} \\
\hline

\multirow{3}{*}{\textcolor{blue}{SF}\text{-UDA}}

& IRG \cite{vs2023instance} & FR50 & \multicolumn{1}{c|}{837} &
\multicolumn{1}{c}{0.0} &
\multicolumn{1}{c}{0.0} &
\multicolumn{1}{c}{\textbf{70.7}} &
\multicolumn{1}{c|}{\underline{33.8}} &
\multicolumn{1}{c}{\underline{26.1}} \\
& LODS \cite{li2022source} & FR101 & \multicolumn{1}{c|}{837} & 
\multicolumn{1}{c}{0.0} & 
\multicolumn{1}{c}{0.0} & 
\multicolumn{1}{c}{3.9} &
\multicolumn{1}{c|}{16.2} &
\multicolumn{1}{c}{5.0} \\
& AASFOD \cite{chu2023adversarial} & FR50 & \multicolumn{1}{c|}{837} & 
\multicolumn{1}{c}{0.11} &
\multicolumn{1}{c}{0.0} &
\multicolumn{1}{c}{54.2} &
\multicolumn{1}{c|}{28.9} &
\multicolumn{1}{c}{20.8} \\
& SF-YOLO \cite{varailhon2024source} & yolov5 & \multicolumn{1}{c|}{837} &
\multicolumn{1}{c}{36.3} & 
\multicolumn{1}{c}{0.0} &
\multicolumn{1}{c}{26.4} &
\multicolumn{1}{c|}{16.8} &
\multicolumn{1}{c}{19.9} \\
\hline
\multirow{1}{*}{\textcolor{blue}{SF}\text{-FSDA}}
&\textbf{AdaptiveCDM}& FR50 & \multicolumn{1}{c|}{\textbf{19}} & 
\multicolumn{1}{c}{\textbf{56.9}} & 
\multicolumn{1}{c}{\textbf{15.2}} & 
\multicolumn{1}{c}{\underline{64.0}} &
\multicolumn{1}{c|}{\textbf{37.4}} &
\multicolumn{1}{c}{\textbf{43.4}} \\
\hline
\end{tabular}
\end{table*}


\setlength{\tabcolsep}{2.2pt}     
\renewcommand{\arraystretch}{1} 

\begin{table*}[t]
\centering
\fontsize{8.0pt}{12.0pt}\selectfont
\caption{Detection comparison on Raabin-WBC. FSDAOD uses complete source data with a few labeled target samples, while AdaptiveCDM uses only k-shot (2,5) target samples without source data. “0” indicates the source-free setting. 
}

\label{tab:raabin_results_few_shot_comparison_with_FSDA}
\begin{tabular}{@{\hspace{1pt}} l 
| l 
| c | c | c c c c c c c c | c c }


\noalign{\hrule height 1pt}
\noalign{\vskip 2pt}   
\multicolumn{14}{c}{Raabin-WBC (M1 $\rightarrow$ M2)} \\
\noalign{\vskip 2pt}   
\hline

\multicolumn{1}{c}{\textbf{}} & {\textbf{Method}} & \textbf{arch.} & \multicolumn{1}{c|}{\textbf{s-imgs}} &
\multicolumn{2}{c|}{\textbf{l-lymph}} &
\multicolumn{2}{c|}{\textbf{neutr.}} &
\multicolumn{2}{c|}{\textbf{s-lymph}} &
\multicolumn{2}{c|}{\textbf{mono.}} &
\multicolumn{2}{c}{\textbf{mAP$_{50}$}} \\
\hline

\multicolumn{4}{c|}{Shots} & \multicolumn{1}{c}{2} & \multicolumn{1}{c|}{5} & 
\multicolumn{1}{c}{2} & \multicolumn{1}{c|}{5} & 
\multicolumn{1}{c}{2} & \multicolumn{1}{c|}{5} & 
\multicolumn{1}{c}{2} & \multicolumn{1}{c|}{5} & 
\multicolumn{1}{c}{2} & \multicolumn{1}{c}{5} \\
\hline

\multirow{4}{*}{\textcolor{blue}{FSDA}} 

& FsDet \cite{wang2020frustratingly} & FR101 & \multicolumn{1}{c|}{2052} &
41.5 & 38.1 &
17.5 & 44.1 &
23.7 & 29.6 &
23.1 & 8.7  &
26.5 & 30.1 \\
& VFA \cite{han2023few} & FR101 & \multicolumn{1}{c|}{2052} &
44.6 & 59.8 &
25.7 & 66.0 &
48.7 & 42.6 &
2.40 & 12.5 &
30.3 & 45.2 \\
& FDP \cite{wang2024fine} & FR101 & \multicolumn{1}{c|}{2052} &
28.2 & 59.1 &
60.7 & 66.8 &
46.2 & 35.2 & 
8.60 & 15.9 &
35.9 & 44.3 \\
& AsyFOD \cite{gao2023asyfod} & yolov5 & \multicolumn{1}{c|}{2052} &
37.2 & 39.3 & 
58.3 & 43.1 &
46.5 & 22.4 & 
0.3 & 0.3  &
35.6 & 26.3 \\
& AcroFOD \cite{gao2022acrofod} & yolov5 & \multicolumn{1}{c|}{2052}  &
37.6 & 59.6 & 
50.5 & 82.1 & 
88.1 & 95.9 &
3.5 & 7.3 &
44.9 & 61.2 \\
& I2DA \cite{inayat2024few} & yolov5 & \multicolumn{1}{c|}{2052}  &
74.1 & 75.2 & 
87.2 & 94.3 &
54.7 & 42.5 &  
40.6 & 71.3 &
\underline{64.2} & \underline{70.8} \\
\hline
\multirow{1}{*}{\text{SF}\textcolor{blue}{-FSDA}}

& \textbf{AdaptiveCDM}& FR50 & \multicolumn{1}{c|}{\textbf{0}} &
22.3 & 51.2 &
75.8 & 81.6 &
93.3 & 93.6 &
73.0 & 75.7 &
\textbf{67.1} & \textbf{75.5} \\
\hline
\end{tabular}

\end{table*}

\renewcommand{\arraystretch}{1}
\setlength{\tabcolsep}{2pt}         

\begin{table*}[t]
\centering
\fontsize{8.0pt}{12.0pt}\selectfont
\caption{Detection comparison on M5 dataset. FSDAOD uses complete source data with a few labeled target samples, while AdaptiveCDM uses only k-shot (2,5) target samples without source data. “0” indicates the source-free setting.}

\label{tab:M5_results_few_shot_comparison_with_FSDA}
\begin{tabular}{@{\hspace{1pt}} l 
| l 
| c | c | c c c c c c c c | c c }

\noalign{\hrule height 1pt}
\noalign{\vskip 2pt}   
\multicolumn{14}{c}{\textbf{M5 (HCM1000x $\rightarrow$ LCM1000x)}}\\
\noalign{\vskip 2pt}   
\hline
 
\multicolumn{1}{c}{} & \textbf{Method} & \textbf{arch.} & \multicolumn{1}{c|}{\textbf{s-imgs}} &
\multicolumn{2}{c|}{\textbf{gamet.}} & 
\multicolumn{2}{c|}{\textbf{schi.}} & 
\multicolumn{2}{c|}{\textbf{troph.}} & 
\multicolumn{2}{c|}{\textbf{ring}} &
\multicolumn{2}{c}{\textbf{mAP$_{50}$}}\\
\hline

\multicolumn{4}{c|}{Shots} & \multicolumn{1}{c}{2} & \multicolumn{1}{c|}{5} & 
\multicolumn{1}{c}{2} & \multicolumn{1}{c|}{5} & 
\multicolumn{1}{c}{2} & \multicolumn{1}{c|}{5} & 
\multicolumn{1}{c}{2} & \multicolumn{1}{c|}{5} & 
\multicolumn{1}{c}{2} & \multicolumn{1}{c}{5} \\
\hline
\multirow{4}{*}{\textcolor{blue}{FSDA}}
& FsDet \cite{wang2020frustratingly} & FR101 & \multicolumn{1}{c|}{837} &
11.8 & 12.3 &
0.0 & 0.0 &
27.7 & 30.7 &
5.4 & 8.6 &
11.2 & 12.9 \\
& VFA \cite{han2023few} & FR101 & \multicolumn{1}{c|}{837}  &
9.1 & 24.5 &
0.1 & 0.2 &
17.8 & 6.5 &
12.8 & 1.3 &
8.9 & 8.1 \\
& FDP \cite{wang2024fine} & FR101 & \multicolumn{1}{c|}{837}  &
16.4 & 33.7 &
1.2 & 1.3 &
27.4 & 33.3 & 
13.5 & 14.0 &
14.6 & 20.5 \\
& AsyFOD \cite{gao2023asyfod} & yolov5 & \multicolumn{1}{c|}{837}  &
14.9 & 36.8 & 
1.2 & 2.8 & 
59.4 & 64.7 & 
28.7 & 30.9 &
26.0 & 33.5 \\
& AcroFOD \cite{gao2022acrofod} & yolov5 & \multicolumn{1}{c|}{837}  &
27.6 & 62.9 & 
17.6 & 5.4 & 
58.7 & 61.3 & 
27.8 & 27.0 &
\underline{32.9} & \underline{39.1} \\
\hline

\multirow{1}{*}{\text{SF}\textcolor{blue}{-FSDA}}
& \textbf{AdaptiveCDM}& FR50 & \multicolumn{1}{c|}{\textbf{0}} &
52.9 & 56.9 &
7.4 & 15.2 &
64.7 & 64.0 &
36.5 & 37.4 &
\textbf{40.4} & \textbf{43.4} \\

\hline
\end{tabular}
\end{table*}

\renewcommand{\arraystretch}{1}
\setlength{\tabcolsep}{15pt}
\begin{table}[t]
\centering
\fontsize{9.0pt}{12.0pt}\selectfont
\caption{Ablation study of the proposed components on M5 
(HCM1000x $\rightarrow$ LCM1000x) under the 5-shot setting.}
\label{tab:ablation}
\begin{tabular}{ccc|c|c}
\hline
$\mathcal{L}_{MT}$ & RAug & CARL & mAP@0.5 & Gain \\
\hline
\checkmark & -- & -- & 36.7 & -- \\
\checkmark & -- & \checkmark & 37.9 & +1.2 \\
\checkmark & \checkmark & -- & 39.2 & +2.5 \\
\checkmark & \checkmark & \checkmark & \textbf{43.4} & \textbf{+6.7} \\
\hline
\end{tabular}
\end{table}


\section{Experiments}
\label{sec: Experiments}
\subsection{Datasets.}
We evaluate our approach on two microscopic benchmark WBC datasets, M5 \cite{sultani2022towards} and Raabin-WBC. 
\textbf{M5} \cite{sultani2022towards} is a large-scale malarial dataset containing images from high-cost (HCM) and low-cost (LCM) microscopes at three resolutions. We selected HCM1000x and LCM1000x images from the source and target domains, respectively. For the source domain, we utilize the original splits provided in \cite{sultani2022towards}.    
\textbf{Raabin-WBC} \cite{kouzehkanan2022large} is a white blood cell dataset collected using two microscopes and annotations from two different doctors. For both datasets, we follow 2,5 shot train splits as per \cite{inayat2024few}.

We also evaluate our methodology on the Realistic-to-Artistic dataset, where the source model is trained on \textbf{Pascal-VOC} and adapted to the target domain, namely \textbf{Clipart}. This dataset contains 1K images and has the same classes as Pascal-VOC. We use the 2-shot and 5-shot splits as given in \cite{inayat2024few}.

\subsection{Implementation Details.}
 We follow the standard source-free setting as in \cite{li2022source,vs2023instance}. We set the number of teacher proposals $N_r$ to 300 and teacher update rate $\eta = 0.9$ in Eq. \ref{eqn:ema}. 
 The source model is fine-tuned on few-shot samples using strong-weak augmentations \cite{vs2023instance} and RAug. 
 In all experiments, the student model is adapted for 10 epochs. To ensure consistency across datasets, loss parameters $\alpha$, $\beta$, and $m$  are set to 1 in Eq. \ref{eqn:CARL}. Final results are reported using the last epoch of the teacher model, with mAP@0.5 averaged over three runs using different random initial seeds, along with standard deviation.

\subsubsection{Data and Code Availability}
The M5, Raabin-WBC, Pascal VOC, and Clipart datasets used in this study are publicly available from their respective sources. The implementation of AdaptiveCDM, along with the few-shot data splits used in our experiments, will be made publicly available upon acceptance to facilitate reproducibility.

\subsection{Results}
In this section, we empirically demonstrate the effectiveness of our proposed method.

\subsubsection{SF-FSDAOD Comparisons:}
In Table \ref{tab:main_results}, AdaptiveCDM improves over the strongest SF-FSDAOD baseline by 4.5 and 6.7 mAP points in the 2-shot and 5-shot settings, respectively.
These improvements can be attributed to the complementary contributions of RAug and CARL; RAug improves target-data diversity and class balance, while CARL enhances foreground–background discrimination and proposal quality.
Notably, AdaptiveCDM achieves 56.9 AP@0.5 for gametocytes and 15.2 AP@0.5 for schizonts in the 5-shot setting, whereas the source model fails to detect either category.

On Raabin-WBC, AdaptiveCDM improves upon the source-only model by 17.2 and 25.6 mAP points in the 2-shot and 5-shot settings, respectively.
AdaptiveCDM also exhibits relatively low variance across three runs, particularly in the 5-shot setting Table ~\ref{tab:main_results}. These results confirm the effectiveness of AdaptiveCDM in low-resource adaptation, consistently improving performance on both common and rare classes.

\subsubsection{Comparison with SF-UDAOD:}
Table~\ref{tab:M5_results_few_shot_comparison_with_SFUDA} compares 
AdaptiveCDM with existing SF-UDAOD methods under different target-supervision 
assumptions. While SF-UDAOD methods utilize the entire unlabeled target dataset, 
AdaptiveCDM relies only on a 5-shot labeled target set without access to source 
data. Therefore, this comparison primarily evaluates target-data efficiency 
rather than performance under identical supervision. Despite using substantially 
fewer target images, AdaptiveCDM achieves higher detection performance than the 
compared SF-UDAOD methods, with margins of at least \textbf{+21.3} mAP points 
on Raabin-WBC and \textbf{+17.3} mAP points on M5. Notably, the compared 
SF-UDAOD methods obtain \textbf{0.0\%} AP@0.5 for the less frequent schizont 
class, whereas AdaptiveCDM achieves improved detection of this category. These 
results suggest that limited but strategically utilized labeled target samples 
can be particularly beneficial for adapting to class-imbalanced microscopic 
datasets.

To further examine cross-domain generalization beyond microscopic imagery, we 
evaluate AdaptiveCDM on Pascal-VOC $\rightarrow$ Clipart, as reported in 
Table~\ref{tab:AdaptiveMCD_clipart}. AdaptiveCDM uses only the 5-shot labeled 
target split, whereas the compared SF-UDAOD methods utilize 500 unlabeled target 
images. Under these different supervision settings, AdaptiveCDM achieves 
\textbf{46.7\%} mAP@0.5, compared with \textbf{45.2\%} for LODS and 
\textbf{31.5\%} for IRG. These results demonstrate that the proposed approach 
generalizes beyond microscopic imagery while maintaining high target-data 
efficiency.

\renewcommand{\arraystretch}{1}
\setlength{\tabcolsep}{12pt}  
\begin{table}[t]
\centering    
\fontsize{8pt}{12.0pt}\selectfont
\caption{Ablation of RAug on M5 (HCM1000x $\rightarrow$ LCM1000x). Comparison with prior augmentation methods under the 5-shot setting. Results report mAP@50(\%) for MT-FreeShot, AdaptiveCDM with CBCP, and our RAug, demonstrating its effectiveness in handling class imbalance and preserving spatial scale.}
\label{tab:ablation_RAug}
\begin{tabular}{c |c |c |c }
\hline
\noalign{\vskip 2pt}   
\textbf{Method} & w/o Augmentation &  Augmentation: CBCP &  Augmentation: RAug \scriptsize{(Ours)}  \\
\noalign{\vskip 2pt}   
\hline
\textbf{mAP$_{50}$} & 36.7 & 40.3 & \textbf{43.4}\\

\hline
\end{tabular}
\end{table}

\begin{figure*}[t]
  \centering
  \includegraphics[width=1
  \linewidth]{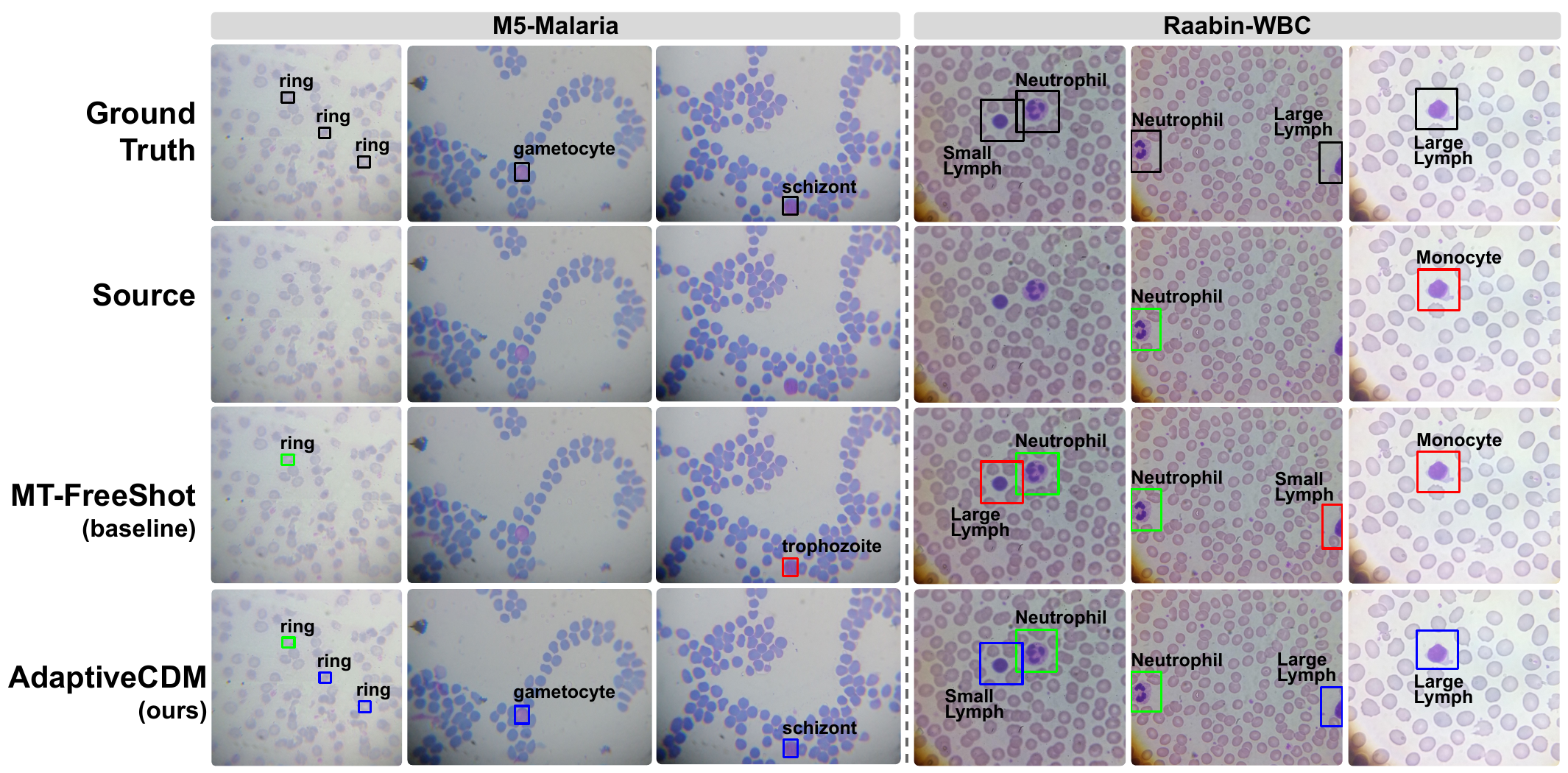}
  \caption{Qualitative detection results on M5 and Raabin-WBC. False positives are shown in \textcolor{red}{red}, correct predictions in \textcolor{green}{green}, and detections missed by the source model and MT-FreeShot but recovered by AdaptiveCDM in \textcolor{blue}{blue}}
  \label{fig:fig_quantitative_results}  
\end{figure*}
\subsubsection{Comparison with FSDAOD: } 
In Table \ref{tab:raabin_results_few_shot_comparison_with_FSDA} and \ref{tab:M5_results_few_shot_comparison_with_FSDA}, 
despite operating without source-domain images, AdaptiveCDM achieves higher overall mAP than the compared FSDAOD methods under the evaluated protocol.
On Raabin-WBC, AdaptiveCDM improves over the strongest compared FSDAOD method by 2.9 and 4.7 mAP points in the 2-shot and 5-shot settings, respectively.
On the M5 dataset, also without access to source-domain images, AdaptiveCDM substantially outperforms by achieving 40.4 and 43.4 mAP@50 in the 2-shot and 5-shot settings respectively.

\renewcommand{\arraystretch}{1}
\setlength{\tabcolsep}{5pt}  
\begin{table*}[t]
\centering
\fontsize{8.0pt}{13.0pt}\selectfont
\caption{Comparison on Pascal-VOC $\rightarrow$ Clipart. Overall mAP@0.5(\%) is reported. 
LODS \cite{li2022source} and IRG \cite{vs2023instance} use full target data, whereas AdaptiveCDM (ours) uses only 5-shot split.
}
\label{tab:AdaptiveMCD_clipart}
\resizebox{\textwidth}{!}{%
\begin{tabular}{c | c | c c | c c | c}
\noalign{\hrule height 1pt}
\noalign{\vskip 1pt}   
\multicolumn{7}{c}{\textbf{Pascal-VOC $\rightarrow$ Clipart}}\\
\noalign{\vskip 1pt}   
\hline
\noalign{\vskip 1pt}
\textbf{Method} &
Source \cite{ren2016faster} &
LODS \cite{li2022source} &
IRG \cite{vs2023instance} &
MT-FreeShot &
AdaptiveCDM &
Oracle \cite{ren2016faster} \\
\noalign{\vskip 1pt}
\hline
\textbf{Target-imgs} & -- & 500 & 500 & 93 & 93 & 500 \\
\textbf{mAP$_{50}$} & 26.9 & 45.2 & 31.5 & 44.9 & \textbf{46.7} & 51.8 \\
\noalign{\hrule height 1pt}
\end{tabular}
}
\end{table*}

\begin{figure*}[t]
  \centering
  \includegraphics[width=1
  \linewidth]{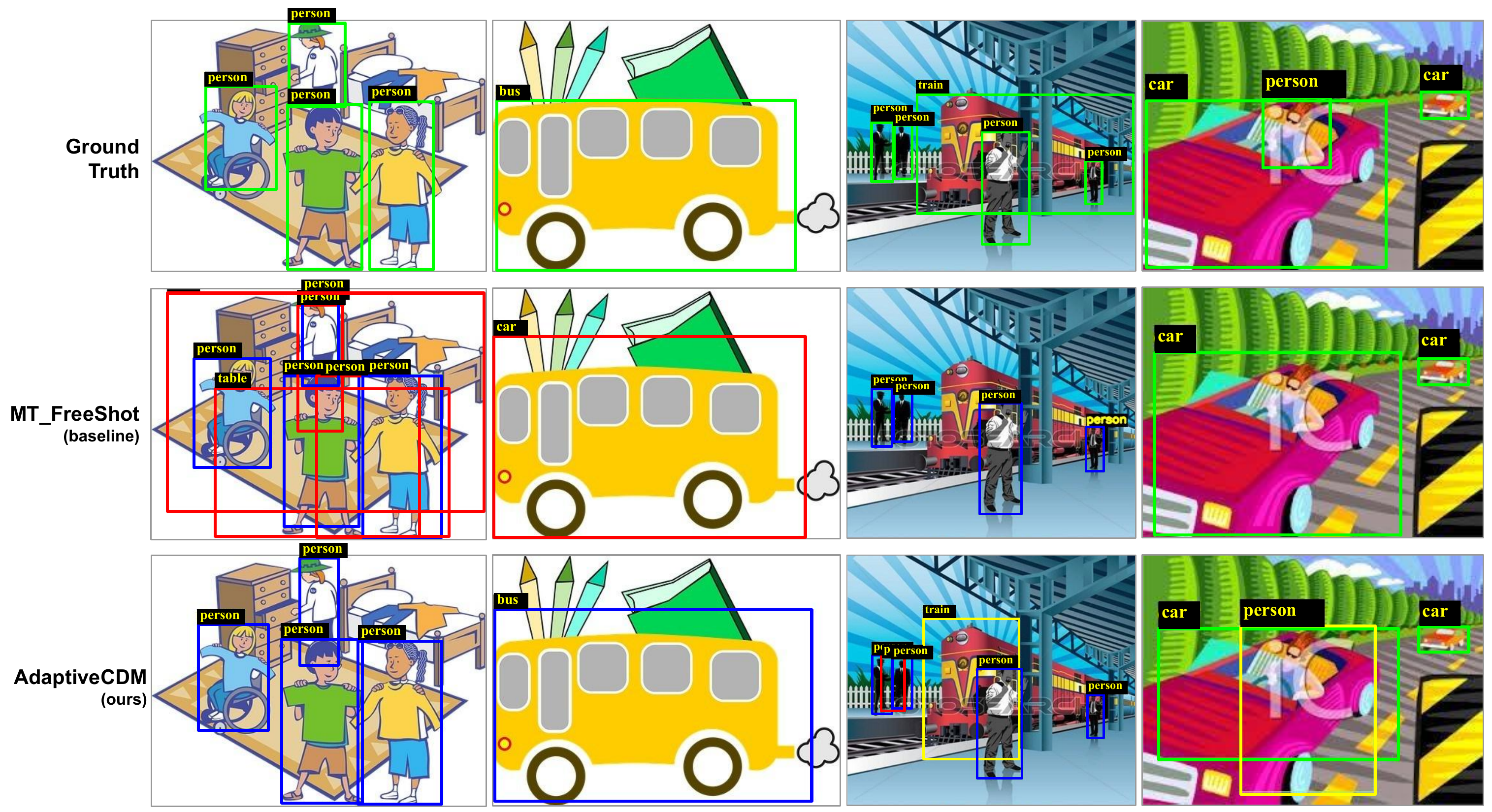}
  \caption{Qualitative results on Pascal-VOC $\rightarrow$ Clipart.  False positives are in \textcolor{red}{red}. Detections missed by baseline model but found by our model are in \textcolor{yellow}{yellow}, highlighting the improved adaptation. Common correct predictions are in \textcolor{blue}{blue} (best visible in color).}
  \label{fig:qualitative_results_clipart}  
\end{figure*}

\subsection{Discussion}
This section provides a detailed analysis of the effectiveness of AdaptiveCDM under the source-free few-shot adaptation setting. We examine how effectively the proposed framework mitigates domain shift with limited target supervision and analyze the individual contributions of RAug and CARL. In particular, we discuss the impact of resolution-aware augmentation on class imbalance and cross-microscope variation, as well as the role of category-aware representation learning in improving feature discrimination and proposal quality.

\subsubsection{Does source-free few-shot adaptation effectively address domain shift?}
The results demonstrate that source-free few-shot adaptation can effectively mitigate domain shift using only a small number of labeled target samples. On M5, the source-only model achieves 15.0 mAP@0.5, indicating a substantial performance degradation when directly transferred from HCM1000x to LCM1000x. Fine-tuning with limited target samples through Faster-FreeShot improves the performance to 35.9 mAP in the 2-shot setting, while MT-FreeShot achieves 34.5 mAP. AdaptiveCDM further increases the performance to \textbf{40.4 }mAP, demonstrating the benefit of explicitly addressing the challenges associated with limited target supervision. In the 5-shot setting, AdaptiveCDM achieves \textbf{43.4} mAP, compared with 31.7 and 36.7 mAP obtained by Faster-FreeShot and MT-FreeShot, respectively.

A similar trend is observed on Raabin-WBC, where AdaptiveCDM achieves \textbf{67.1/75.5} mAP@0.5 under \textbf{2-/5-shot} adaptation, despite having no access to source-domain images during adaptation. Moreover, AdaptiveCDM achieves competitive or superior performance compared with several FSDAOD methods that retain access to the complete source dataset. These findings indicate that a small amount of strategically utilized labeled target data can provide an effective alternative when source data are inaccessible, particularly in privacy-restricted biomedical applications.

\subsubsection{Why does RAug improve adaptation?}
RAug addresses three key challenges in few-shot microscopic imagery: limited target data, class imbalance, and resolution variation. Unlike conventional augmentation, RAug generates instance-balanced samples while preserving the spatial scale and morphology of cellular structures. 
The ablation results in Table \ref{tab:ablation_RAug} support its effectiveness: performance improves from 36.7 mAP@0.5 without augmentation to 40.3 mAP@0.5 with CBCP and 43.4 mAP@0.5 with RAug on M5 under the 5-shot setting.

These results indicate that preserving resolution-aware spatial characteristics while balancing rare classes improves cross-microscope adaptation.

\subsubsection{Why does CARL improve detection?}
CARL improves foreground-background discrimination and class-level feature separability, which are particularly important when different cell categories exhibit similar visual characteristics. Using teacher-guided spatial cues, CARL encourages intra-class compactness and inter-class separation, leading to more discriminative representations and improved proposal quality. The ablation results in Table \ref{tab:ablation} show that adding CARL to the model improves performance from 39.2 to 43.4 mAP@0.5, while the qualitative results show fewer erroneous background proposals. Thus, RAug and CARL play complementary roles: RAug improves target-data diversity and balance, while CARL enhances feature discrimination and localization.

\subsection{Ablation} 
Table~\ref{tab:ablation} analyzes the contribution of each component of AdaptiveCDM on M5 under the 5-shot setting. The MT-FreeShot baseline achieves 36.7 mAP@0.5. Incorporating CARL improves the performance to 37.9 mAP@0.5, demonstrating the benefit of category-aware representation learning for improving feature discrimination. RAug provides a larger individual improvement, increasing performance to 39.2 mAP@0.5, indicating the importance of addressing class imbalance while preserving resolution-dependent spatial characteristics. Combining RAug and CARL further improves performance to 43.4 mAP@0.5, corresponding to an overall gain of 6.7 mAP points over the baseline. These results indicate that RAug and CARL provide complementary benefits, with RAug improving target-data diversity and class balance while CARL enhances discriminative representation learning.

\subsection{Qualitative Results}
Figure \ref{fig:fig_quantitative_results} presents a qualitative comparison of AdaptiveCDM with the source model and MT-FreeShot baseline. On M5, the source and baseline models struggle to distinguish visually similar malarial stages and frequently miss rare classes under domain shift. In contrast, AdaptiveCDM improves foreground--background discrimination and detection of visually similar and underrepresented cell categories, resulting in fewer missed detections and false positives. Similarly, on Raabin-WBC, AdaptiveCDM better distinguishes morphologically similar cell classes and improves detection under varying illumination conditions, particularly for Neutrophils and Large Lymphocytes. 
Figure \ref{fig:qualitative_results_clipart} further presents qualitative results for Pascal-VOC $\rightarrow$ Clipart adaptation. Compared with MT-FreeShot, AdaptiveCDM produces more accurate detections with fewer missed objects and false positives, demonstrating that the proposed approach generalizes beyond microscopic imagery to a challenging realistic-to-artistic domain shift.

\section{Limitations}
\label{sec: limitations}
Our study has limitations that provide directions for future research. The experiments primarily focus on two microscopic imaging datasets and assume a shared label space between the source and target domains. The performance of AdaptiveCDM may also depend on the quality of the source-pretrained detector. Furthermore, CARL depends partly on teacher-generated spatial cues; although these are combined with ground-truth annotations, inaccurate teacher proposals under severe domain shifts may introduce noisy supervision. These observations suggest that future work could investigate uncertainty-aware proposal selection, stronger rare-class modeling, and adaptation across more severe variations in microscopes, magnifications, and acquisition conditions.

\section{Conclusion}
\label{sec: conclusion}
We introduce AdaptiveCDM, a framework for Source-Free Few-Shot Domain Adaptive Object Detection (SF-FSDAOD) in microscopic imagery, where a pre-trained source model is adapted using only a limited number of labeled target samples without accessing the source data. To mitigate domain discrepancies, we propose Resolution-Aware Augmentation (RAug), which preserves spatial scale fidelity while alleviating class imbalance. To enhance instance-level feature alignment and proposal quality, we introduce Category-Aware Representation Learning (CARL). Extensive experiments demonstrate the effectiveness of AdaptiveCDM under limited target supervision.
The proposed method achieves 40.4/43.4 mAP@0.5 on M5 and 67.1/75.5 mAP@0.5 on Raabin-WBC under 2-/5-shot adaptation, respectively, while demonstrating strong generalization on Pascal-VOC$\rightarrow$Clipart. 
Ablation experiments further demonstrate that RAug and CARL provide complementary benefits, jointly improving the M5 5-shot performance from 36.7 to 43.4 mAP@0.5.
These results highlight its potential for adapting microscopic cell detectors across imaging devices and acquisition settings where sharing source-domain data is impractical.

\section{Declaration of generative AI and AI-assisted technologies}
During the preparation of this work, the authors used ChatGPT (OpenAI) solely to assist with textual refinement, including improving language, clarity, readability, grammar, and sentence structure. The tool was not used to generate research data, perform experiments, develop the study methodology, or formulate the scientific conclusions of the work. After using this tool, the authors reviewed and edited the content as needed and take full responsibility for the content of the published article.


 \bibliographystyle{elsarticle-num} 
 \bibliography{ref}







\end{document}